%% file: main.tex
\documentclass{article} 
\usepackage{iclr2027_conference,times}
\usepackage{amsmath} 
\usepackage{amssymb}  
\usepackage{mathtools}
\usepackage{graphicx}
\usepackage{subcaption}
\input{math_commands.tex}

\usepackage{multirow}
\usepackage{booktabs}
\usepackage{hyperref}
\usepackage{xcolor}
\usepackage{url}
\usepackage{setspace}
\usepackage{wrapfig}
\usepackage{enumitem}

\newcommand{\best}[1]{\textbf{#1}}

\AtBeginDocument{\setlength{\abovedisplayskip}{4pt}\setlength{\belowdisplayskip}{4pt}\setlength{\abovedisplayshortskip}{2pt}\setlength{\belowdisplayshortskip}{2pt}}

\title{Align Then Reason: A Multimodal Lip-Sync Judge for Dubbing}

\author{
Rui Liu\thanks{Work done during an internship at Netflix.}~~$^{1}$,
Bhavin Jawade$^{2}$, 
Haoqi Li$^{2}$, 
Shivam Mehta$^{2}$, 
Karan Saxena$^{2}$, \\  
\textbf{Yinghong Lan}$^{2}$, 
\textbf{Cameron R. Wolfe}$^{2}$ \\
$^{1}$University of Maryland, College Park, 
$^{2}$Netflix \\
}

\newcommand{\ours}{\textsc{atr}}

\iclrfinalcopy 
\begin{document}

\maketitle

\vspace{-6pt}
\begin{abstract}

Dubbing quality control requires a reference-free judge that can determine whether a candidate text line matches a speaker's visible articulation in both content and timing, using only silent video and text because dubbed audio may not yet exist. Existing visual speech recognizers and video-language models are poorly suited to this setting: even when fine-tuned to recover spoken content from lip motion, they remain largely insensitive to temporal errors. We introduce \textit{Align Then Reason} (\ours), a multilingual lip-sync judge that first establishes a monotonic alignment between frame-level lip representations and the phonetic units of the candidate line, then reasons over this alignment to make the final judgment. An alignment scorer provides the LLM with both local evidence for each phonetic unit and a calibrated global alignment score, enabling it to reason jointly about content and timing. On a seven-language benchmark, our method improves mean AUC over the corresponding Qwen3.5 SFT baselines by $59.4\%$, $50.2\%$, and $50.8\%$ with 2B, 4B, and 9B reasoners, respectively. The gains generalize across LLM families, reaching mean AUC improvements of $45.9\%$ and $46.6\%$ over the best baseline for LLaMA-3.1-8B and Mistral-7B, respectively. They also transfer across datasets to three unseen MuAViC languages. Furthermore, we evaluate on two downstream tasks built from real dubbing lines. On dub-line reranking, \ours-9B outperforms the best lip-reading baseline by $52.0\%$, while on script-to-clip assignment, \ours-9B improves over the best lip-reading baseline by $17.7\%$. 



\end{abstract}


\input{intro}
\input{approach}

\input{exp}
\input{related}
\input{con}

\bibliography{ref}
\bibliographystyle{plainnat}
\input{appendix}

\end{document}

%% file: math_commands.tex
\usepackage{amsmath,amsfonts,bm}

\def\eqref#1{equation~\ref{#1}}

\def\1{\bm{1}}

\DeclareMathAlphabet{\mathsfit}{\encodingdefault}{\sfdefault}{m}{sl}
\SetMathAlphabet{\mathsfit}{bold}{\encodingdefault}{\sfdefault}{bx}{n}



%% file: intro.tex
\section{Introduction}
\label{sec:introduction}

Dubbing \citep{federico2020dubbing, brannon2023dubbing, chaume2020audiovisual} increasingly relies on generative systems to translate, rewrite, and synthesize speech at scale. Yet producing a linguistically plausible translated line is not sufficient: the line must also agree with the visible articulation of the speaker. A poor dub may preserve the intended meaning but begin too early, lag behind the mouth, or contain phonetic content that is incompatible with the observed lip motion. Detecting such failures manually is costly and difficult to scale, motivating a lip-sync judge that can score candidate lines before speech synthesis.

This setting is more constrained than conventional audio-visual synchronization \citep{korbar2018cooperative, iashin2024synchformer, javed2025audio}. At review time, the candidate dubbed audio may not yet exist, so the judge needs to operate from only a silent video $V$ and a candidate text line $c$. A useful judge must therefore answer a cross-modal question: \emph{does this candidate text agree with what the mouth appears to be saying, and does that agreement unfold in the correct temporal order?} This requires sensitivity to two complementary failure modes, including content and temporal. Content errors occur when the candidate line is incompatible with the visible articulation, whereas temporal errors occur when otherwise compatible speech is misaligned or presented in an incorrect temporal order. The judge should additionally generalize across languages, as for dubbing, we may need to translate to many different languages.

However, achieving such a lip-sync judge is challenging. Previous visual speech recognition models \citep{shi2022learning, ma2023autoavsr, yeo2024vspllm, cappellazzo2025large} are designed to infer a transcript from visual speech rather than score an arbitrary candidate line against a video. Audio-visual synchronization models  \citep{chung2016out, javed2025audio} explicitly measure temporal correspondence between mouth motion and speech, but require an audio stream and do not condition on candidate text. Large video-language models \citep{bai2025qwen3} can in principle condition jointly on video and text, but temporal reasoning is not guaranteed by multimodal pretraining alone \citep{liu2024tempcompass, li2025temporal, qi2026improve}. Prior analyses have shown that strong performance on common video-language benchmarks can often be obtained with weak or even absent temporal modeling, revealing substantial atemporal and single-frame biases \citep{buch2022revisiting, lei2023revealing}. These limitations are especially consequential for lip-sync judgment, where the candidate text can remain identical while only the ordering or timing of the mouth motion changes. Our experiments show that this problem persists when existing models are employed as lip-sync judges. When we hold the candidate text fixed and reverse, shift or freeze the video frames, video-language models (e.g., Qwen3.5 \citep{bai2025qwen3}) from 2B to 9B remain close to chance. These results indicate that recognizing visual or linguistic content does not by itself provide the fine-grained temporal correspondence required for lip-sync quality control.

\begin{figure}[t]
\centering
\includegraphics[width=\linewidth]{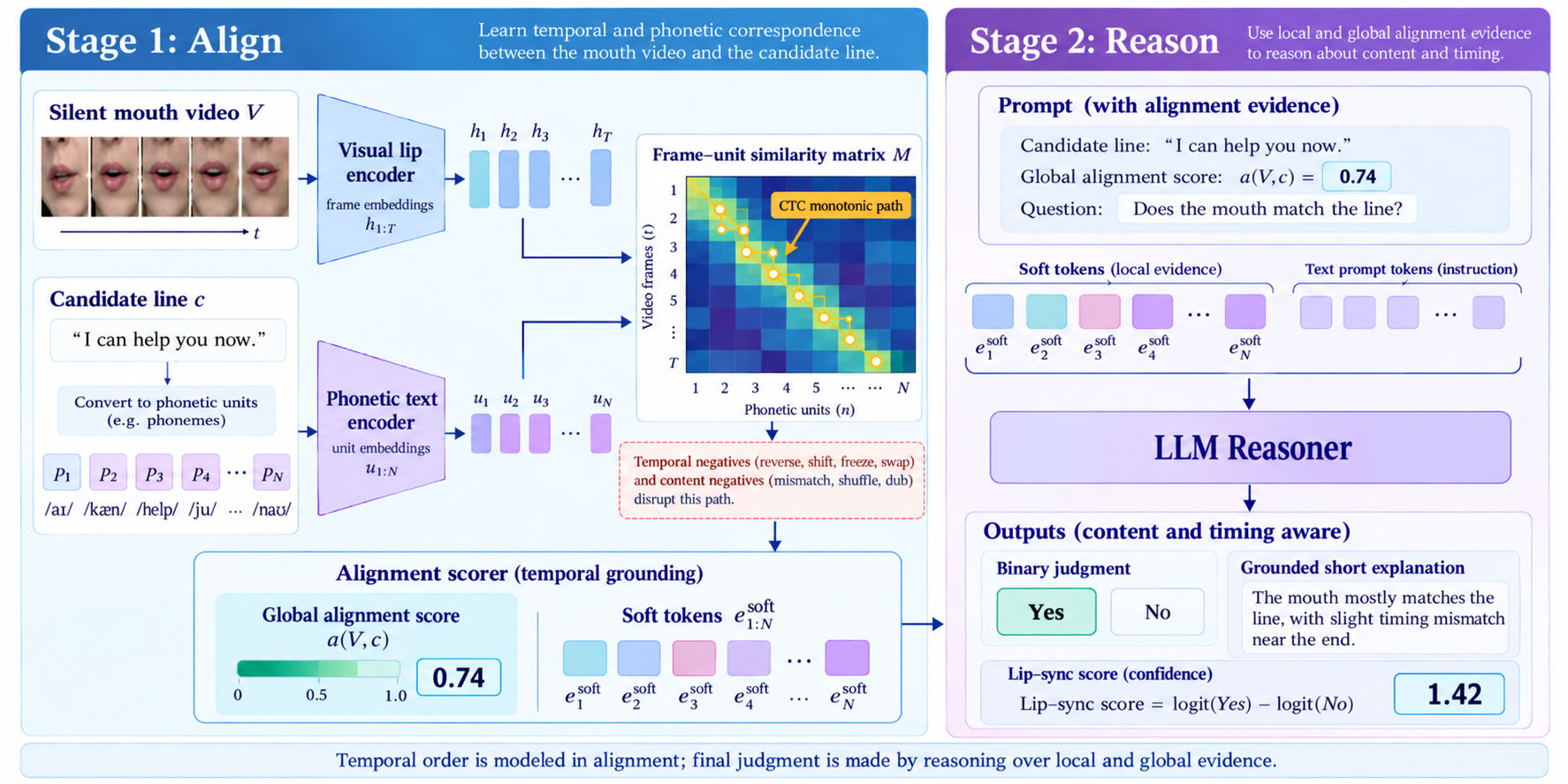}
\caption{\textbf{Overview of Align Then Reason (\ours).} \emph{Stage 1 (Align):} the silent mouth video is encoded into frame embeddings $\mathbf{h}_{1:T}$ and the candidate line into phonetic-unit embeddings $\mathbf{u}_{1:N}$. Their similarity $M_{t,n}$ is scored by CTC whose labels are the \emph{positions} of the units in the line, so only a monotonic path through $M$ (highlighted) scores well, and reordering, shifting, or freezing the frames breaks it. Trained with content and temporal negatives, the scorer emits a calibrated global score $a(V,c)$ and per-unit soft tokens $\mathbf{e}^{\mathrm{soft}}_{1:N}$. \emph{Stage 2 (Reason):} the soft tokens are prepended to a prompt containing $a(V,c)$ and the line, and an LLM answers whether the mouth matches the line; the difference of its \texttt{Yes} and \texttt{No} logits is the lip-sync score.}
\label{fig:overview}
\vspace{-5pt}
\end{figure}


To address these limitations, we propose Align Then Reason (ATR). We first formulate lip-sync judgment as monotonic cross-modal alignment between frame-level visual lip representations and phonetic representations of the candidate line. Given a video and candidate text, we convert the text into a sequence of phonetic units and construct a frame-by-unit similarity matrix. A compatible pair should admit a strong monotonic path through this matrix, whereas content substitutions and temporal perturbations disrupt that path. We score this structure using a candidate-dependent CTC \citep{graves2006connectionist} formulation and train the alignment model with complementary content and temporal negatives. Temporal order is therefore encoded directly in the scoring structure. By representing text phonetically rather than with a language-specific output vocabulary, the same alignment mechanism can also be generalized across languages. A single alignment score, however, necessarily compresses the evidence from the entire utterance. We therefore introduce an LLM reasoner on top of the alignment scorer. For each candidate line, the scorer provides two complementary forms of evidence: per-phonetic-unit soft tokens that summarize local visual-phonetic correspondence and a calibrated scalar that summarizes global whole-line monotonic alignment. The soft tokens are supplied directly as continuous embeddings before the ordinary text-token embeddings, while the calibrated scalar is included in the textual prompt. The reasoner is trained on genuine and incorrect candidate lines paired with the same video and produces both a binary judgment and a short grounded explanation. Its \texttt{Yes} and \texttt{No} logits difference is used as the final lip-sync score.

We evaluate the proposed judge along complementary dimensions designed to separate content discrimination, temporal sensitivity, multilingual generalization, and downstream usefulness. Across a seven-language benchmark, our method boosts mean AUC over corresponding Qwen3.5 SFT baselines by $59.4\%$, $50.2\%$, and $50.8\%$ at the 2B, 4B, and 9B reasoner scales, respectively. These performance gains generalize across different LLM families, outperforming the top baseline by $45.9\%$ on LLaMA-3.1-8B \citep{grattafiori2024llama} and $46.6\%$ on Mistral-7B \citep{jiang2023mistral}, while also successfully transferring to three unseen MuAViC languages \citep{anwar2023muavic}. Finally, on two downstream applications constructed from real dubbing lines, \ours-9B surpasses the strongest lip-reading baseline by $52.0\%$ on dub-line reranking and $17.7\%$ on script-to-clip assignment.  Together, these results validate a effective judge: explicit monotonic alignment provides strong temporal grounding, while the LLM reasoner improves content discrimination and produces an interpretable final judgment. In summary, our key contributions are as follows:
\begin{itemize}[left=0pt]
    \item We identify and measure a structural gap between existing models and reference-free text--video lip-sync judgment: autoregressive visual-speech and video-language models are insensitive to temporal orders.
    
    \item We propose a temporally grounded multilingual lip-sync judge that decouples temporal tracking from semantic evaluation. By coupling a candidate-conditioned CTC scorer to an LLM via per-unit soft tokens and a calibrated global score, we directly anchor LLM reasoning in fine-grained alignment dynamics.
    
    \item Across seven languages, multiple LLM backbones, and unseen datasets, our approach improves mean AUC by up to $59.4\%$ over baselines and outperforms the top lip-reading models on downstream dubbing tasks by up to $52.0\%$.
\end{itemize}


%% file: approach.tex
\vspace{-6pt}
\section{Approach}
\label{sec:approach}

\vspace{-5pt}
\subsection{Problem Formulation}
\vspace{-5pt}
We consider dubbing lip-sync quality evaluation from a silent video and a candidate text line. Given a video $V=\{v_t\}_{t=1}^{T}$ and a candidate line $c$, our goal is to estimate how well the content and timing of $c$ agree with the observed mouth movements. No dubbed audio and no reference transcript are available at inference time. Our lip-sync judge operates in two stages (Figure~\ref{fig:overview}). First, a cross-modal monotonic alignment scorer learns to align visual lip motion with the phonetic units of a candidate line and produces local and global alignment evidence. Second, an LLM reasoner consumes the candidate text together with this alignment evidence and predicts whether the mouth movements match the line.


\vspace{-6pt}
\subsection{Cross-Modal Monotonic Alignment}
\label{sec:cross_modal_alignment}

\vspace{-6pt}
Lip-sync quality requires both content agreement and temporal consistency: the observed articulations should correspond to the phonetic content of the candidate line and occur in the same order. We therefore formulate the first stage as monotonic alignment between video frames and candidate-dependent phonetic units. Given $V$ and $c$, a visual lip encoder produces frame-level embeddings $\mathbf{h}_{1:T}$, with $\mathbf{h}_t\in\mathbb{R}^{d}$, while a phonetic text encoder converts $c$ into a sequence of $N$ phonetic units and produces embeddings $\mathbf{u}_{1:N}$, with $\mathbf{u}_n\in\mathbb{R}^{d}$. Their frame-to-unit compatibility is measured by cosine similarity,
$M_{t,n}=\mathbf{h}_t^\top\mathbf{u}_n/(\|\mathbf{h}_t\|\|\mathbf{u}_n\|)$.

\vspace{-6pt}
\paragraph{Monotonic alignment scorer.}
Some frames carry no speech (pauses, closed mouth, transitions), so we add a blank state with a learned scalar logit $b$. The emission logits are $E_{t,0}=b$ for the blank and $E_{t,n}=M_{t,n}$ for $n=1,\ldots,N$. A softmax over $k=0,\ldots,N$ yields the frame-level emission probabilities $p_{t,k}$, the probability that frame $t$ corresponds to text unit $k$.

We use Connectionist Temporal Classification (CTC) \citep{graves2006connectionist} to enforce monotonicity. Unlike conventional CTC, whose labels correspond to entries in a fixed character or phoneme vocabulary, our alignment labels are positions within the current candidate line: label $n$ denotes the $n$-th phonetic unit of $c$. The target sequence is therefore $\mathbf{y}_c=(1,2,\ldots,N)$. We define the alignment score as the length-normalized CTC log-likelihood
\begin{equation}
s_{\mathrm{align}}(V,c)
=
\frac{1}{N}
\log
\sum_{\boldsymbol{\pi}\in\mathcal{B}^{-1}(\mathbf{y}_c)}
\prod_{t=1}^{T}
p_{t,\pi_t},
\label{eq:ctc_score}
\end{equation}
where $\boldsymbol{\pi}$ is a frame-level CTC path and $\mathcal{B}$ is the standard CTC collapse operator. This construction allows a phonetic unit to span multiple consecutive frames and permits blank frames between units, while requiring the candidate units to appear in their original order. Consequently, the score reflects both phonetic compatibility and temporal progression.

\paragraph{Contrastive alignment training.}
We train the two encoders such that a genuine video-text
pair receives a higher alignment score than corrupted alternatives.  For a genuine pair, let $s^{+}=s_{\mathrm{align}}(V,c^{+})$, and let $s_x^{-}$ denote the score of a negative example of type $x$. We use the smooth margin loss $h(\Delta)=\log(1+\exp(1-\Delta))$, where $\Delta=s^{+}-s_x^{-}$. Two families of negatives target the two failure modes. \emph{Content negatives} $\mathcal{C}=\{\texttt{mismatch},\texttt{shuffle},\texttt{dub}\}$ keep the video and change the text: another clip's line of similar length, the true line with its words permuted. The dub translation serves as a natural content negative. These negatives encourage sensitivity to linguistic correspondence. \emph{Temporal negatives} $\mathcal{R}=\{\texttt{reverse},\texttt{shift},\texttt{freeze},\texttt{swap}\}$ keep the text and corrupt the video: reversing the frames, circularly shifting them, holding one frame for a span, or exchanging the halves, encouraging the model to distinguish correct temporal alignment from visually similar but temporally inconsistent sequences. The alignment-training objective is
\begin{equation}
\begin{aligned}
\mathcal{L}_{\mathrm{align}}
={}&
\frac{1}{|\mathcal{C}|}
\sum_{x\in\mathcal{C}}
\mathbb{E}\!\left[
h\!\left(s^{+}-s_x^{-}\right)
\right]
+
\frac{1}{|\mathcal{R}|}
\sum_{x\in\mathcal{R}}
\mathbb{E}\!\left[
h\!\left(s^{+}-s_x^{-}\right)
\right]
+ \mathcal{L}_{\mathrm{aux}},
\end{aligned}
\label{eq:alignment_objective}
\end{equation}


\vspace{-6pt}
\paragraph{Auxiliary phonetic supervision.}
The contrastive objective constrains pairwise alignment scores but does not directly require the visual embeddings to preserve fine-grained phonetic information. We therefore attach a phoneme-recognition head to the same frame embeddings: a linear map $A$ produces logits over a global phoneme vocabulary plus blank, and given the phoneme sequence $\mathbf{z}$ of the true transcript, the auxiliary loss is standard CTC, $\mathcal{L}_{\mathrm{aux}}=-\log\sum_{\boldsymbol{\pi}\in\mathcal{B}^{-1}(\mathbf{z})}\prod_{t=1}^{T}\operatorname{softmax}(A\mathbf{h}_t)_{\pi_t}$. The phoneme targets come from the true transcript and are used only during training, so, as in auxiliary modality learning \citep{liu2025mmcd, liu2026caml}, they shape the visual representation without being needed at inference.


\vspace{-6pt}
\subsection{LLM Reasoner}
\label{sec:llm_reasoner}
The alignment score is a strong structured signal, but it reduces the whole line to a single number. This makes it hard to tell why a line fails or to distinguish the correct line from another well-aligned but incorrect one. We therefore use a language model that takes the candidate text together with two views of the scorer's evidence: per-unit representations that preserve local alignment information and a calibrated global score for the whole clip.

\vspace{-6pt}
\paragraph{Local evidence: soft tokens.}
For each candidate unit $n$ we summarize the frames that best match it. From the similarity matrix $M$ we compute frame-normalized attention weights $\alpha_{t,n}=\exp(M_{t,n})/\sum_{t'}\exp(M_{t',n})$, the visual summary $\bar{\mathbf{h}}_n=\sum_{t}\alpha_{t,n}\mathbf{h}_t$, and the strongest local match $\rho_n=\max_t M_{t,n}$. The evidence for unit $n$ is $\mathbf{x}_n=[\mathbf{u}_n \mid \bar{\mathbf{h}}_n \mid \rho_n]\in\mathbb{R}^{2d+1}$, where $\mathbf{u}_n$ says which sound is being tested, $\bar{\mathbf{h}}_n$ summarizes the mouth frames aligned to it, and $\rho_n$ says how well its best frame matches. A trainable projection with layer normalization maps each $\mathbf{x}_n$ to the language model's hidden size, $\mathbf{e}^{\mathrm{soft}}_n=\operatorname{LN}(P\mathbf{x}_n+\mathbf{b}_P)\in\mathbb{R}^{d_{\mathrm{LM}}}$. We call these \emph{soft tokens}: they occupy token positions but are vectors produced by the scorer, not vocabulary entries. They are concatenated in front of the prompt's token embeddings, $\mathbf{E}_{\mathrm{in}}=[\mathbf{e}^{\mathrm{soft}}_{1:N},E_{\mathrm{LM}}(\omega_{1:M})]$, and processed as one sequence.


\vspace{-6pt}
\paragraph{Global evidence: calibrated scalar.}
The soft tokens carry local matches; $s_{\mathrm{align}}(V,c)$ carries the globally normalized monotonic fit of the whole line. To put it on a stable scale we estimate its mean $\mu_s$ and standard deviation $\sigma_s$ over genuine pairs in a held-out calibration split and define the calibrated scalar $a(V,c)=\operatorname{sigmoid}\big((s_{\mathrm{align}}(V,c)-\mu_s)/\sigma_s\big)$, which is written into the prompt as a two-decimal number (template in Appendix~\ref{app:prompt}).


\vspace{-6pt}
\paragraph{Reasoner training.}
Training examples are grouped by video so that each genuine line is contrasted with wrong lines for the \emph{same} mouth sequence: for each training video $V_i$ we form the positive $(V_i,c_i^{+})$ and $K$ content negatives $\{(V_i,c_{i,j}^{-})\}_{j=1}^{K}$. The positive is supervised with \texttt{Yes} and each negative with \texttt{No}, followed by a short explanation naming the part of the line (early, middle, or late) that contains the weakest unit $n^{*}=\arg\min_n\rho_n$, or stating that the match is consistent throughout. Let $e^{+}$ and $\{e_j^{-}\}$ be the reasoner inputs of the group. Teacher-forced language-model cross-entropy $\mathcal{L}_{\mathrm{LM}}(e)$ supervises both the binary \texttt{Yes}/\texttt{No} decision and the corresponding grounded explanation with the following loss: 
$
\mathcal{L}_{\mathrm{reasoner}}
=
\frac{1}{1+K}
\sum_{e\in\{e^{+},e_1^{-},\ldots,e_K^{-}\}}
\mathcal{L}_{\mathrm{LM}}(e).
$

\vspace{-6pt}
\paragraph{Final score.}
At inference time, the complete lip-sync judge applies the two trained components sequentially to an input pair $(V,c)$. The frozen alignment scorer first computes the candidate-specific soft tokens $\{\mathbf{e}^{\mathrm{soft}}_n\}_{n=1}^{N}$ and the calibrated global alignment score $a(V,c)$. These signals are combined with the candidate-text prompt and passed to the LLM reasoner. The final lip-sync score is the logit difference between the \texttt{Yes} and \texttt{No} answer tokens:
$
s_{\mathrm{judge}}(V,c)
=
\ell_{\mathrm{Y}}(V,c)
-
\ell_{\mathrm{N}}(V,c)
=
\log
\frac{
P(\mathrm{Yes}\mid V,c)
}{
P(\mathrm{No}\mid V,c)
}.
\label{eq:judge_score}
$
A larger $s_{\mathrm{judge}}(V,c)$ indicates stronger agreement between the candidate line and the observed mouth movements.

%% file: exp.tex
\vspace{-6pt}
\section{Experiments}
\label{sec:experiments}

\vspace{-5pt}
\subsection{Experimental Setup}
\label{sec:setup}



\vspace{-6pt}
\paragraph{Data construction.} \label{sec:data}

We construct a dataset from on-screen dialogue segments, where each sample consists of a talking-face video, a source-language transcript, and a dub line. The dataset contains approximately 50K multilingual training samples and 2,100 test samples. The test benchmark is balanced across seven languages including English, Spanish, French, Japanese, Korean, Brazilian Portuguese, and Turkish.

For each sample, we construct two content negatives: mismatch (a length-matched source line from another sample in the same language), a word-order shuffle. The dub translation serves as a natural content negative. We also generate four temporal negatives by reversing, circularly shifting, freezing, or swapping halves of the visual frames while keeping the text fixed. We use Auto-AVSR \citep{ma2023autoavsr} as the visual encoder and XPhoneBERT \citep{nguyen2023xphonebert} as the text encoder. Additional architectural and training details are provided in Appendix~\ref{app:setup}.

\vspace{-6pt}
\paragraph{Baselines and evaluation.}
We compare against a broad set of baselines spanning visual speech recognition and vision-language models. For visual speech recognition, we employ several models as judges: Auto-AVSR \citep{ma2023autoavsr} scores each candidate using its CTC log-likelihood, while the autoregressive models AV-HuBERT \citep{shi2022learning}, LLaMA-AVSR \citep{cappellazzo2025large}, and VSP-LLM \citep{yeo2024vspllm} use the candidate's conditional log-likelihood under the decoder. We also evaluate Qwen3.5 vision-language models at 2B, 4B, and 9B scales \citep{bai2025qwen3}, both zero-shot and after SFT to predict the spoken line from the video. To test whether our method generalizes across reasoners, we further evaluate \ours{} with different LLM backbones, including Qwen3.5, LLaMA-3.1-8B \citep{grattafiori2024llama}, and Mistral-7B \citep{jiang2023mistral}.

Our primary evaluation metric is pooled AUC for each corruption axis. Genuine video-text pairs are treated as positives and their corrupted counterparts as negatives; AUC measures how well the model separates genuine from corrupted pairs across the full test set. We additionally report paired accuracy, which measures whether each genuine pair scores higher than its corresponding corruption.

\vspace{-6pt}
\subsection{Main Results}
\label{sec:main_results}
\vspace{-5pt}
We evaluate performance on the seven-language benchmark across seven distinct corruption types, categorized into content corruptions (mismatch, shuffle, dub) and temporal corruptions (reverse, shift, freeze, swap), as defined in Section \ref{sec:data}. Performance is measured by pooled AUC (higher is better). In the absence of reference answers, AUC quantifies whether the judge correctly assigns a higher alignment score to genuine positive pairs than to corrupted negative pairs.

As shown in Table~\ref{tab:main}, existing baselines exhibit a pronounced performance gap compared to \ours{} across both content and temporal axes. Among the baseline models, Qwen3.5 benefits substantially SFT on content corruptions: scaling from 2B to 9B parameters improves its AUC on mismatch, shuffle, and dub to $0.730$, $0.737$, and $0.743$, respectively. However, its performance on temporal corruptions remains near chance level, with AUCs across all four temporal axes hovering between $0.510$ and $0.520$. Autoregressive speech recognizers suffer from a similar temporal blind spot. While Auto-AVSR captures some temporal signal ($0.582$--$0.670$ AUC), it still falls far short of \ours{}. In English-only evaluations (Table~\ref{tab:english}, Appendix~\ref{app:english}), Auto-AVSR achieves competitive mismatch performance, but \ours-9B substantially outperforms it and all other baselines across the remaining six evaluation axes.

\begin{table}[ht]
\caption{\textbf{Performance on the seven-language benchmark measured by pooled AUC ($\uparrow$).} Existing speech recognition and vision-language baselines struggle on temporal corruptions, staying near chance ($0.50$). In contrast, \ours{} achieves balanced, high accuracy across all seven evaluation axes and multiple LLM backbones, reaching a mean AUC of $0.920$ with Qwen3.5-9B, a $50.8\%$ relative improvement over the best baseline ($0.610$).}
\vspace{-5pt}
\label{tab:main}
\begin{center}
\resizebox{\linewidth}{!}{
\begin{tabular}{lcccccccc}
\toprule
& \multicolumn{3}{c}{Content} & \multicolumn{4}{c}{Temporal} & \\
\cmidrule(lr){2-4} \cmidrule(lr){5-8}
Method & Mismatch & Shuffle & Dub & Reverse & Shift & Freeze & Swap & Mean \\
\midrule
Auto-AVSR      & 0.602 & 0.599 & 0.507 & 0.670 & 0.582 & 0.609 & 0.620 & 0.598 \\
AV-HuBERT          & 0.500 & 0.578 & 0.520 & 0.513 & 0.506 & 0.501 & 0.508 & 0.518 \\
LLaMA-AVSR        & 0.472 & 0.793 & 0.484 & 0.531 & 0.513 & 0.500 & 0.521 & 0.545 \\
VSP-LLM          & 0.481 & 0.764 & 0.508 & 0.524 & 0.507 & 0.511 & 0.517 & 0.545 \\
Qwen3.5-2B Base    & 0.515 & 0.448 & 0.570 & 0.503 & 0.499 & 0.503 & 0.501 & 0.506 \\
Qwen3.5-4B Base    & 0.527 & 0.477 & 0.586 & 0.500 & 0.498 & 0.500 & 0.501 & 0.513 \\
Qwen3.5-9B Base    & 0.537 & 0.479 & 0.605 & 0.502 & 0.500 & 0.503 & 0.502 & 0.518 \\
Qwen3.5-2B SFT    & 0.632 & 0.604 & 0.685 & 0.502 & 0.501 & 0.502 & 0.503 & 0.561 \\
Qwen3.5-4B SFT    & 0.696 & 0.739 & 0.750 & 0.510 & 0.505 & 0.505 & 0.510 & 0.602 \\
Qwen3.5-9B SFT    & 0.730 & 0.737 & 0.743 & 0.520 & 0.513 & 0.510 & 0.519 & 0.610 \\
\midrule
\ours~(Qwen3.5-2B)         & 0.782 & 0.967 & 0.821 & 0.945 & 0.909 & 0.905 & 0.932 & 0.894 \\
\ours~(Qwen3.5-4B)          & 0.784 & 0.960 & 0.841 & 0.955 & 0.926 & 0.916 & 0.944 & 0.904 \\
\ours~(Qwen3.5-9B)         & \textbf{0.786} & 0.974 & \textbf{0.862} & \textbf{0.964} & \textbf{0.931} & \textbf{0.970} & \textbf{0.952} & \textbf{0.920} \\
\ours~(LLaMA-3.1-8B) & 0.771 & \textbf{0.976} & 0.812 & 0.924 & 0.900 & 0.931 & 0.918 & 0.890 \\
\ours~(Mistral-7B) & 0.767 & 0.969 & 0.836 & 0.956 & 0.914 & 0.872 & 0.945 & 0.894 \\
\bottomrule
\end{tabular}
}
\end{center}
\vspace{-8pt}
\end{table}

In contrast, \ours{} demonstrates consistently strong discrimination across all seven axes and across diverse LLM backbones. When paired with Qwen3.5-9B, \ours{} achieves $0.786$, $0.974$, and $0.862$ AUC on the three content corruptions, and $0.964$, $0.931$, $0.970$, and $0.952$ AUC on the four temporal corruptions, yielding a leading overall mean AUC of $0.920$. Compared to the strongest baseline mean ($0.610$), this represents a $50.8\%$ relative improvement ($+0.310$ absolute AUC). These performance gains are robust across different LLM backbones: \ours{} maintains mean AUCs between $0.890$ and $0.920$ across Qwen3.5, LLaMA-3.1-8B, and Mistral-7B. Multi-seed experiments (Appendix~\ref{app:seeds}) further confirm the statistical stability of these gains.

\begin{table}[ht]
\caption{\textbf{Paired accuracy ($\uparrow$) on the seven-language benchmark.} Baseline speech recognizers and frontier models degrade significantly under temporal corruptions, demonstrating insensitivity to fine-grained timing. \ours{} consistently achieves superior accuracy across all content and temporal axes regardless of the underlying LLM backbone.}
\vspace{-5pt}
\label{tab:paired}
\begin{center}
\resizebox{\linewidth}{!}{
\begin{tabular}{lcccccccc}
\toprule
& \multicolumn{3}{c}{Content} & \multicolumn{4}{c}{Temporal} & \\
\cmidrule(lr){2-4} \cmidrule(lr){5-8}
Method & Mismatch & Shuffle & Dub & Reverse & Shift & Freeze & Swap & Mean \\
\midrule
Auto-AVSR & 0.624 & 0.751 & 0.509 & 0.956 & 0.892 & 0.710 & 0.897 & 0.763 \\
AV-HuBERT & 0.496 & 0.639 & 0.519 & 0.524 & 0.541 & 0.514 & 0.519 & 0.536 \\
LLaMA-AVSR & 0.471 & 0.874 & 0.496 & 0.607 & 0.596 & 0.463 & 0.612 & 0.588 \\
VSP-LLM & 0.480 & 0.900 & 0.534 & 0.597 & 0.534 & 0.518 & 0.585 & 0.593 \\
Qwen3.5-9B Base & 0.553 & 0.456 & 0.595 & 0.522 & 0.499 & 0.552 & 0.499 & 0.525 \\
Qwen3.5-9B SFT & 0.752 & 0.799 & 0.780 & 0.601 & 0.595 & 0.565 & 0.623 & 0.674 \\
\midrule
Claude Opus 5 & 0.588 & \textbf{0.990} & 0.676 & 0.485 & 0.731 & 0.946 & 0.740 & 0.737 \\
Gemini 3.1 Pro & 0.561 & 0.973 & 0.681 & 0.580 & 0.598 & 0.690 & 0.749 & 0.690 \\
GPT-5.6-terra & 0.654 & 0.900 & 0.592 & 0.540 & 0.603 & 0.695 & 0.716 & 0.671 \\
GPT-5.6-luna & 0.481 & 0.880 & 0.578 & 0.540 & 0.728 & 0.724 & 0.662 & 0.656 \\
GPT-5.6-sol & 0.613 & 0.931 & 0.632 & 0.616 & 0.739 & 0.852 & 0.694 & 0.725 \\
\midrule
\ours~(Qwen3.5-2B) & 0.778 & 0.973 & 0.827 & 0.977 & 0.980 & 0.917 & 0.978 & 0.919 \\
\ours~(Qwen3.5-4B) & 0.768 & 0.971 & 0.847 & 0.980 & 0.984 & 0.929 & 0.984 & 0.923 \\
\ours~(Qwen3.5-9B) & \textbf{0.781} & 0.978 & \textbf{0.864} & \textbf{0.989} & \textbf{0.988} & \textbf{0.987} & 0.986 & \textbf{0.939} \\
\ours~(LLaMA-3.1-8B) & 0.771 & 0.983 & 0.812 & 0.965 & 0.969 & 0.946 & 0.965 & 0.916 \\
\ours~(Mistral-7B) & 0.777 & 0.980 & 0.840 & 0.995 & 0.994 & 0.905 & \textbf{0.994} & 0.926 \\

\bottomrule
\end{tabular}
}
\end{center}
\vspace{-8pt}
\end{table}

To complement the pooled AUC metric, Table~\ref{tab:paired} reports paired accuracy across the seven-language macro-average. We additionally evaluate state-of-the-art frontier models, including Claude Opus 5 \citep{anthropic_claude_opus_5_2026}, Gemini 3.1 Pro \citep{gemini31pro2026}, and GPT-5.6 variants \citep{openai2026gpt56}. While frontier models demonstrate strong capabilities on coarse content corruptions (e.g., reaching up to $0.990$ accuracy on shuffle), their performance drops sharply on temporal corruptions, averaging between $0.656$ and $0.737$ overall. In contrast, \ours{} maintains paired accuracies exceeding $0.916$ across all model variants, highlighting its superior temporal sensitivity.

\vspace{-6pt}
\subsection{Generalization Testing}
\label{sec:muavic}
\vspace{-6pt}
\paragraph{Generalization across datasets and languages.}

 
To test cross-dataset and cross-lingual generalization, we evaluate on the MuAViC dataset \citep{anwar2023muavic} across German, Arabic, and Russian, three languages absent from our training set. The evaluation protocol incorporates the same corruption suite, excluding dub as MuAViC lacks dubbing tracks. Table~\ref{tab:muavic} summarizes the results. Auto-AVSR fails on content corruptions for Arabic and Russian ($0.546$--$0.552$ AUC). Meanwhile, the Qwen3.5-9B Base model remains near chance ($0.501$--$0.502$ AUC) across all temporal corruptions. Without retraining on MuAViC, \ours-9B outperforms Qwen3.5-9B Base by $+0.237$ mean AUC ($+46.1\%$ improvement). As described in Section \ref{sec:approach}, the reasoner ingests a score derived from globally normalized monotonic alignment, calibrated on genuine pairs. When evaluating out-of-domain data, domain shifts in raw scalar distributions can cause genuine pairs to fall outside the expected decision bounds. To address this without model fine-tuning, we refit only two scalar normalization parameters ($\mu_s$ and $\sigma_s$) on a held-out set of 300 target-language genuine samples, requiring no parameter updates. This recalibration restores sensitivity across all temporal axes, notably raising freeze performance from $0.572$ to $0.937$ AUC. Overall, recalibrated \ours-9B reaches a mean AUC of $0.875$, representing a $+0.361$ absolute gain ($+70.2\%$ improvement) over the Qwen3.5-9B Base model and a $+0.297$ absolute gain ($+51.4\%$ improvement) over Auto-AVSR.

\begin{table}[ht]
\caption{\textbf{Generalization on MuAViC across three unseen languages (German, Arabic, and Russian).} No model weights are updated. Out-of-the-box \ours-9B achieves $0.751$ mean AUC. Adjusting the two normalization constants ($\mu_s, \sigma_s$) for recalibration lifts the mean AUC to $0.875$, outperforming Auto-AVSR by $+51.4\%$ relatively.}
\vspace{-5pt}
\label{tab:muavic}
\begin{center}
\resizebox{\linewidth}{!}{
\begin{tabular}{lccccccc}
\toprule
Method & Mismatch & Shuffle & Reverse & Shift & Freeze & Swap & Mean \\
\midrule
Auto-AVSR & 0.546 & 0.552 & 0.635 & 0.555 & 0.601 & 0.580 & 0.578 \\
Qwen3.5-9B Base & 0.580 & 0.498 & 0.501 & 0.501 & 0.502 & 0.501 & 0.514 \\
\ours-9B & 0.661 & 0.930 & 0.790 & 0.763 & 0.572 & 0.787 & 0.751 \\
\ours-9B (recalibrated) & \textbf{0.707} & \textbf{0.946} & \textbf{0.927} & \textbf{0.827} & \textbf{0.937} & \textbf{0.905} & \textbf{0.875} \\
\bottomrule
\end{tabular}
}
\end{center}
\vspace{-10pt}
\end{table}

\paragraph{Generalization across LLM backbones.}

To evaluate flexibility across underlying architectures, we benchmark \ours{} across diverse open-source LLMs. Beyond the Qwen3.5 family, \ours{} transfers effectively to LLaMA-3.1-8B ($0.890$ mean AUC, $0.916$ paired accuracy) and Mistral-7B ($0.894$ mean AUC, $0.926$ paired accuracy), as shown in Tables \ref{tab:main} and \ref{tab:paired}. These consistent gains demonstrate that the proposed reasoning framework generalizes robustly across different LLM architectures.

\vspace{-6pt}
\subsection{Ablation Studies}
\label{sec:ablations}

In this section, we conduct ablation studies to evaluate the individual contributions of our core design choices, including auxiliary phonetic supervision, the LLM reasoner, soft tokens, and the calibrated scalar. The results are summarized in Table~\ref{tab:ablation}.

\vspace{-6pt}
\paragraph{Effects of auxiliary phonetic supervision.}
The contrastive alignment objective constrains relative pair scores but does not explicitly require the visual representation to preserve fine-grained phonetic information. Auxiliary phonetic supervision provides this additional constraint by encouraging the lip features to retain distinctions between speech sounds. Removing it reduces the mean AUC from $0.920$ to $0.898$, with the largest degradation on the content axes, confirming that phonetic supervision primarily strengthens content discrimination.

\vspace{-6pt}
\paragraph{Effects of the LLM reasoner.}
While the alignment scorer alone effectively captures temporal alignment, it struggles significantly with content comprehension. Introducing the LLM reasoner substantially improves performance on the content-oriented metrics while preserving the model's strong temporal reasoning capabilities.

\vspace{-6pt}
\paragraph{Effects of soft tokens.}
Removing the soft tokens lowers mean AUC from $0.920$ to $0.872$, particularly on content mismatches. The soft tokens provide local alignment evidence for individual phonetic units, complementing the global scalar that summarizes the overall monotonic fit. Together, the two signals provide the reasoner with both local evidence about \emph{which} units match and global evidence about \emph{how well} the full sequence aligns.

\vspace{-6pt}
\paragraph{Effects of the calibrated scalar.}
Removing the calibrated global score causes the largest degradation, reducing mean AUC from $0.920$ to $0.672$ and sharply weakening temporal discrimination. This follows from the model design: the video-dependent components of each soft token pool over frames and therefore do not explicitly preserve temporal order. In contrast, the global alignment score is computed from the monotonic alignment path and directly captures sequence order. Removing this signal therefore drives performance on \texttt{reverse}, \texttt{shift}, and \texttt{swap} close to chance. \texttt{Freeze} remains more detectable because it changes the frame content itself rather than only reordering frames.

\begin{table}[ht]
\caption{\textbf{Ablations on core components of \ours~(Macro AUC $\uparrow$).} Phonetic supervision and the LLM reasoner are critical for content understanding, while soft tokens provide local alignment evidence. The calibrated scalar is essential for temporal sensitivity; removing it causes the largest degradation (reducing mean AUC to 0.672). The full model achieves the best overall performance.}
\vspace{-3pt}
\label{tab:ablation}
\begin{center}
\resizebox{\linewidth}{!}{
\begin{tabular}{l ccc cccc c}
\toprule
& \multicolumn{3}{c}{Content} & \multicolumn{4}{c}{Temporal} & \\
\cmidrule(lr){2-4} \cmidrule(lr){5-8}
Method & Mismatch & Shuffle & Dub & Reverse & Shift & Freeze & Swap & Mean \\
\midrule
\ours-9B & 0.786 & 0.974 & 0.862 & 0.964 & 0.931 & 0.970 & 0.952 & 0.920 \\
\quad w/o phonetic supervision & 0.751 & 0.944 & 0.793 & 0.961 & 0.921 & 0.965 & 0.949 & 0.898 \\
\quad w/o LLM reasoner & 0.695 & 0.757 & 0.701 & 0.970 & 0.935 & 0.971 & 0.962 & 0.856 \\
\quad w/o soft tokens & 0.683 & 0.932 & 0.748 & 0.953 & 0.905 & 0.942 & 0.940 & 0.872 \\
\quad w/o calibrated scalar & 0.671 & 0.927 & 0.724 & 0.525 & 0.534 & 0.805 & 0.517 & 0.672 \\
\bottomrule
\end{tabular}
}
\vspace{-8pt}
\end{center}
\end{table}

\vspace{-6pt}
\subsection{Downstream Evaluation}
\label{sec:downstream}

\vspace{-6pt}
We further evaluate whether the learned alignment signal transfers beyond synthetic corruptions to two downstream dubbing tasks built from real candidate lines: \emph{dub-line reranking} and \emph{script-to-clip assignment}. The first tests whether the model can distinguish subtle phonetic differences among semantically equivalent lines, while the second tests whether it can recover the correct correspondence between real script lines and video clips.

\begin{wraptable}{r}{0.5\linewidth}
\vspace{-10pt}
\caption{\textbf{Dub-line reranking.} Each clip is paired with its professional dub line and three meaning-preserving, length-matched paraphrases; the model must rank the professional line first. The alignment scorer achieves $0.479$ Top-1 accuracy, outperforming the strongest lip-reading baseline ($0.316$) by $52\%$ relative. MRR denotes mean reciprocal rank.}
\label{tab:rerank}
\centering
\setlength{\tabcolsep}{9pt}
\begin{tabular}{lcc}
\toprule
Method & Top-1$\uparrow$ & MRR$\uparrow$ \\
\toprule
Chance & 0.250 & 0.521 \\
Isochrony (length fit) & 0.237 & 0.494 \\
Auto-AVSR & 0.315 & 0.558 \\
AV-HuBERT & 0.244 & 0.510 \\
LLaMA-AVSR & 0.316 & 0.575 \\
VSP-LLM & 0.309 & 0.572 \\
\ours-9B & \best{0.479} & \best{0.684} \\
\bottomrule
\end{tabular}
\vspace{-12pt}
\end{wraptable}

\vspace{-6pt}
\paragraph{Dub-line reranking.}
For each of the 2100 test samples, we pair the professional dub line with three LLM-generated paraphrases that preserve its meaning and remain within $\pm15\%$ of its length. The model scores all four candidates and must rank the professional line first, making chance Top-1 accuracy $0.250$. This is the selection step of a pipeline that samples several candidate lines, with the judge acting as a verifier, analogous to test-time scaling for LLMs, which samples multiple candidates and decides how to allocate and aggregate them \citep{zheng2026parallelprobe, dai2026small, zheng2026llms}. As shown in Table~\ref{tab:rerank}, the \ours-9B achieves $0.479$ Top-1 accuracy, compared with $0.316$ for the strongest lip-reading baseline, a $52\%$ relative improvement. 


\vspace{-6pt}
\paragraph{Script-to-clip assignment.}
We next evaluate whether the judge can recover which real script line belongs to which shot. For the in-domain evaluation, we construct five-way assignment problems by sampling five clips from the same title and language and shuffling their five professional dub lines. The judge scores all $25$ clip--line pairs, after which we solve a maximum-weight one-to-one matching so that each line is assigned exactly once. We report both per-clip accuracy and Exact Block accuracy, which requires all five assignments in a block to be correct; chance performance is $0.200$ and $0.008$, respectively. As shown in Table~\ref{tab:assign}, \ours-9B achieves $0.746$ per-clip accuracy and $0.522$ Exact Block accuracy, outperforming Auto-AVSR by $17.7\%$ relative in per-clip accuracy and by $0.139$ absolute in Exact Block accuracy. It also exceeds the strong length-matching heuristic by $0.090$ absolute in per-clip accuracy.

For MuAViC, the evaluation is naturally constructed from pairs of clips and candidate lines, yielding a two-way assignment problem rather than the five-way blocks used in-domain. Under one-to-one matching with two items, assignments are either both correct or both swapped, making per-clip and exact-block accuracy equivalent; we therefore report only per-clip accuracy. \ours~(Qwen3.5-9B) achieves 0.704 compared with 0.598 for Auto-AVSR, an improvement of 17.7\%. Unlike dub-line reranking, where fine-grained phonetic alignment is most useful, script-to-clip assignment benefits more from the full reasoner because candidate lines differ substantially in content, allowing it to combine local alignment evidence with global semantic consistency.


\begin{wraptable}{r}{0.6\textwidth}
\centering
\vspace{-10pt}
\caption{\textbf{Script-to-clip assignment.} In-domain, we form five-clip blocks from the same title and language and match their five shuffled professional lines one-to-one; Per-clip measures individual assignments, while Exact Block requires all five to be correct. \ours-9B achieves $0.746$ per-clip accuracy, improving over Auto-AVSR by $17.7\%$. On MuAViC, where the available evaluation naturally forms two-way clip--line pairs, \ours~reaches $0.704$ versus $0.598$ for Auto-AVSR.}
\label{tab:assign}
\vspace{-3pt}
\setlength{\tabcolsep}{5pt}
\resizebox{\linewidth}{!}{%
\begin{tabular}{lccc}
\toprule
& \multicolumn{2}{c}{In-domain (5-way)}
& MuAViC (2-way) \\
\cmidrule(lr){2-3}
\cmidrule(lr){4-4}
Method & Per-clip$\uparrow$ & Exact Block$\uparrow$ & Per-clip$\uparrow$ \\
\midrule
Chance & 0.200 & 0.008 & 0.500 \\
Isochrony (length fit) & 0.656 & 0.397 & 0.500 \\
Auto-AVSR & 0.634 & 0.383 & 0.598 \\
AV-HuBERT & 0.365 & 0.081 & 0.548 \\
LLaMA-AVSR & 0.396 & 0.134 & 0.571 \\
VSP-LLM & 0.354 & 0.096 & 0.556 \\
\ours-2B & 0.724 & 0.483 & 0.683 \\
\ours-9B & \best{0.746} & \best{0.522} & \best{0.704} \\
\bottomrule
\end{tabular}%
}
\vspace{-13pt}
\end{wraptable}

%% file: related.tex
\vspace{-6pt}
\section{Related Work}
\label{sec:related_work}

\vspace{-6pt}
\paragraph{Dubbing.}
Prior dubbing systems mainly enforce audiovisual coherence through duration, prosody, and isochrony constraints \citep{federico2020dubbing,tam2022isochrony}. These signals capture timing structure but not whether a candidate line is phonetically compatible with the observed articulation. In contrast, we score candidate text directly against silent lip motion before audio synthesis, using multilingual phonetic representations \citep{nguyen2023xphonebert} to support a shared judge across languages.

\vspace{-6pt}
\paragraph{Visual speech recognition.}
Visual speech recognition models recover text from lip motion through audio--visual representation learning, large-scale transcription, or LLM-based decoding \citep{shi2022learning,ma2023autoavsr,yeo2024vspllm,cappellazzo2025large}. Lip-sync metrics based on SyncNet instead align mouth motion with audio \citep{chung2016out,prajwal2020wav2lip}, while visual forced alignment localizes a known transcript in silent video \citep{kim2023dvfa}. None directly addresses reference-free scoring of an arbitrary candidate line from silent video: recognizers optimize transcription, synchronization methods require audio, and forced alignment assumes frame-level localization targets. Our method instead learns candidate-conditioned visual--phonetic alignment without alignment labels, using monotonic CTC to score both content compatibility and temporal order.

\vspace{-6pt}

\paragraph{LLMs as judges.}
LLMs and vision-language models are increasingly used as learned evaluators and reward models \citep{zheng2023judging, liu2026deltarubric}, and non-textual evidence can be injected into a language model through continuous prefix representations \citep{tsimpoukelli2021frozen, lester2021prompt}. However, video-language models often exhibit weak temporal reasoning \citep{buch2022revisiting, lei2023revealing, liu2024tempcompass, li2025temporal}, and multimodal reasoning remains brittle when the visual input is degraded or perceptually ambiguous \citep{liu2026reinforcing, liu2025dual}. We therefore do not rely on the LLM to infer temporal correspondence directly from video. Instead, a monotonic alignment scorer provides structured local evidence and a calibrated global score, while the LLM performs higher-level content reasoning over these signals. This separates temporal grounding from semantic judgment, in the spirit of decoupling visual perception from language reasoning \citep{li2025self}, and makes the calibration of the scorer's output part of the interface between the two stages \citep{guo2017calibration}, much as uncertainty-aware guidance weights a teacher signal by its calibrated reliability under domain shift \citep{liu2026adaptive}.

%% file: con.tex
\vspace{-6pt}
\section{Conclusions}
\label{sec:con}
\vspace{-6pt}

We introduce Align Then Reason, a reference-free multilingual lip-sync judge designed to evaluate whether candidate text matches a speaker's visual lip movements in both content and timing without requiring reference audio. By coupling frame-level visual phonetic alignment with LLM reasoning, our approach overcomes the temporal insensitivity in video-language models and script limitations in standard visual speech recognizers. Extensive evaluations across a seven-language benchmark demonstrate significant performance gains across diverse model families, achieving up to a 50.8\% mean AUC improvement with a 9B reasoner, alongside strong zero-shot generalization to unseen languages. Crucially, its superior performance on downstream tasks, outperforming top lip-reading baselines by 52.0\% on dub-line reranking and 17.7\% on script-to-clip assignment, highlights the effectiveness of our approach. Our reasoner is trained with supervised fine-tuning; reinforcement learning, which has recently strengthened reasoning in LLMs and multimodal models \citep{zheng2026parallel, dai2026cde, liu2026save, liu2026stable}, is a natural next step for training the judge. More broadly, systems that improve themselves without human review depend on a reliable automatic evaluator \citep{zheng2026llms, zheng2026dream}, and \ours~could serve as that signal for rewriting dub lines.

%% file: appendix.tex
\newpage
\appendix


\section{Experimental Setup}
\label{app:setup}
Given a video, we use Auto-AVSR \citep{ma2023autoavsr} as the visual encoder to produce a 768-d feature for each frame. A temporal lip adapter then maps these features into the alignment space using layer normalization and a linear projection to 512 dimensions, followed by a two-layer temporal Transformer with eight attention heads and a 2048-dimensional feed-forward layer. A final layer normalization and linear projection yield a 256-dimensional lip embedding $\ell_t\in\mathbb{R}^{256}$ for each frame $t$.

For text, we convert each candidate line into phonetic units and encode them with XPhoneBERT \citep{nguyen2023xphonebert}, producing 768-d representations. A linear projection maps them into the same 256-dimensional alignment space. We adapt the LLM reasoner with LoRA \citep{hu2021lora}, using rank 16 and $\alpha=32$. All models are trained with AdamW \citep{loshchilov2017decoupled} using a learning rate of $3\times10^{-4}$, weight decay of $0.01$, and batch size 32.

\section{Reasoner Prompt}
\label{app:prompt}
The soft tokens $\mathbf{e}^{\mathrm{soft}}_{1:N}$ are prepended to the embeddings of the following prompt, in which $a(V,c)$ is rendered with two decimals:
\begin{quote}
\ttfamily
Alignment confidence: $a(V,c)$.\\
Candidate line: $c$\\
Question: Do the mouth movements match the candidate line?\\
Answer:
\end{quote}
During training the answer is \texttt{Yes} or \texttt{No} followed by the grounded explanation; at inference we read the logits of the two answer tokens at the answer position.

\section{English-Only and Per-Language Results}
\label{app:english}

Because speech recognition baselines such as Auto-AVSR were trained primarily on English speech corpora, we report English-only performance separately in Table~\ref{tab:english}. On English data, Auto-AVSR serves as a competitive content baseline, achieving the highest performance on \texttt{mismatch} ($0.784$ AUC). However, \ours-9B substantially outperforms Auto-AVSR and all other baselines across the remaining six evaluation axes, achieving strong results on content corruptions (\texttt{shuffle} $0.989$, \texttt{dub} $0.971$) and temporal corruptions ($0.940$--$0.979$ AUC).

Table~\ref{tab:perlang} breaks down performance across individual languages for three representative corruptions. These results highlight why baseline performance degrades on the full seven-language macro-benchmark: Auto-AVSR drops to chance level ($0.50$ AUC) on Japanese (\texttt{ja}) and Korean (\texttt{ko}) due to tokenizer and vocabulary constraints on non-Latin scripts, while its temporal discrimination remains weak across non-English languages ($0.57$--$0.67$ AUC). In contrast, our alignment scorer demonstrates robust cross-lingual generalizability, maintaining strong content discrimination ($0.63$--$0.78$ AUC) and near-ceiling temporal sensitivity ($\ge 0.96$ AUC) across all seven languages.

\begin{table}[ht]
\caption{\textbf{English-only evaluation measured by pooled AUC ($\uparrow$).} Recognizer baselines were trained primarily on English data. While Auto-AVSR achieves strong content performance on English (winning \texttt{mismatch}), \ours-9B outperforms all baselines across the remaining six content and temporal axes.}
\label{tab:english}
\begin{center}
\resizebox{0.9\linewidth}{!}{
\begin{tabular}{lccccccc}
\toprule
& \multicolumn{3}{c}{Content} & \multicolumn{4}{c}{Temporal} \\
\cmidrule(lr){2-4} \cmidrule(lr){5-8}
Method & Mismatch & Shuffle & Dub & Reverse & Shift & Freeze & Swap \\
\midrule
Auto-AVSR & \textbf{0.784} & 0.817 & 0.946 & 0.896 & 0.639 & 0.714 & 0.825 \\
AV-HuBERT & 0.544 & 0.864 & 0.964 & 0.584 & 0.512 & 0.527 & 0.545 \\
LLaMA-AVSR & 0.587 & 0.930 & 0.505 & 0.599 & 0.531 & 0.521 & 0.546 \\
VSP-LLM & 0.562 & 0.927 & 0.477 & 0.598 & 0.533 & 0.549 & 0.575 \\
Qwen3.5-9B Base & 0.483 & 0.535 & 0.737 & 0.502 & 0.502 & 0.506 & 0.498 \\
Qwen3.5-9B SFT & 0.756 & 0.787 & 0.740 & 0.524 & 0.511 & 0.512 & 0.514 \\
\midrule
\ours-9B & 0.765 & \textbf{0.989} & \textbf{0.971} & \textbf{0.979} & \textbf{0.940} & \textbf{0.958} & \textbf{0.973} \\
\bottomrule
\end{tabular}
}
\end{center}
\end{table}

\begin{table}[ht]
\caption{\textbf{Per-language AUC comparison between Auto-AVSR and our alignment scorer on three representative axes.} Auto-AVSR degrades to chance level on non-Latin scripts (Japanese and Korean) due to tokenization constraints and shows weak temporal sensitivity on non-English languages. Our alignment scorer maintains consistent effectiveness across all seven languages, remaining near ceiling ($\ge 0.96$ AUC) on temporal corruptions.}
\label{tab:perlang}
\begin{center}
\resizebox{0.9\linewidth}{!}{
\begin{tabular}{lcccccc}
\toprule
& \multicolumn{2}{c}{Mismatch} & \multicolumn{2}{c}{Reverse} & \multicolumn{2}{c}{Freeze} \\
\cmidrule(lr){2-3} \cmidrule(lr){4-5} \cmidrule(lr){6-7}
Language & Auto-AVSR & Scorer & Auto-AVSR & Scorer & Auto-AVSR & Scorer \\
\midrule
en & \textbf{0.78} & 0.63 & 0.90 & \textbf{0.98} & 0.71 & \textbf{0.96} \\
es & 0.60 & \textbf{0.72} & 0.65 & \textbf{0.96} & 0.65 & \textbf{0.97} \\
fr & 0.57 & \textbf{0.67} & 0.64 & \textbf{0.97} & 0.64 & \textbf{0.97} \\
ja & 0.50 & \textbf{0.73} & 0.57 & \textbf{0.97} & 0.46 & \textbf{0.99} \\
ko & 0.50 & \textbf{0.78} & 0.60 & \textbf{0.98} & 0.54 & \textbf{0.97} \\
pt-BR & 0.61 & \textbf{0.70} & 0.67 & \textbf{0.97} & 0.64 & \textbf{0.97} \\
tr & \textbf{0.65} & 0.64 & 0.66 & \textbf{0.96} & 0.62 & \textbf{0.97} \\
\bottomrule
\end{tabular}
}
\end{center}
\end{table}

\section{Statistical Significance}
\label{app:seeds}

To evaluate statistical significance of our results, we train all LLM reasoner backbones across three different random seeds. Table~\ref{tab:seeds} reports the mean AUC and standard deviation for the seven-language benchmark, zero-shot MuAViC, and recalibrated MuAViC. 

Across all reasoner architectures, \ours{} demonstrates consistently low variance on the seven-language benchmark (e.g., $0.910 \pm 0.015$ for Qwen3.5-9B and $0.895 \pm 0.012$ for LLaMA-3.1-8B). On the out-of-domain MuAViC evaluation, initial uncalibrated scores exhibit higher variance due to domain shifts in scalar distributions ($0.778 \pm 0.060$ for Qwen3.5-9B). Normalizing the scalar via two-parameter recalibration reduces variance while significantly boosting mean accuracy ($0.852 \pm 0.033$). These results confirm that performance gains are statistically sound and robust across random initializations.

\begin{table}[ht]
\caption{\textbf{Multi-seed evaluation reporting mean AUC and standard deviation across three random training seeds.} Performance gains remain stable across diverse LLM reasoner backbones on the 7-language benchmark, out-of-domain MuAViC dataset, and recalibrated MuAViC evaluation.}
\vspace{-3pt}
\label{tab:seeds}
\begin{center}
\resizebox{\linewidth}{!}{
\begin{tabular}{lccc}
\toprule
Model Backbone & Seven-Language Macro & MuAViC (Zero-Shot) & MuAViC (Recalibrated) \\
\midrule
\ours~(Qwen3.5-2B) & $0.884 \pm 0.018$ & $0.746 \pm 0.074$ & $0.815 \pm 0.041$ \\
\ours~(Qwen3.5-4B) & $0.895 \pm 0.023$ & $0.748 \pm 0.070$ & $0.802 \pm 0.073$ \\
\ours~(Qwen3.5-9B) & $0.910 \pm 0.015$ & $0.778 \pm 0.060$ & $0.852 \pm 0.033$ \\
\ours~(LLaMA-3.1-8B) & $0.895 \pm 0.012$ & $0.711 \pm 0.018$ & $0.832 \pm 0.011$ \\
\ours~(Mistral-7B) & $0.900 \pm 0.020$ & $0.777 \pm 0.085$ & $0.821 \pm 0.045$ \\
\bottomrule
\end{tabular}
}
\end{center}
\end{table}